\documentclass{article}
\usepackage{spconf,amsmath,graphicx}
\usepackage[export]{adjustbox}
\usepackage{url}
\usepackage{algorithm}
\usepackage[noend]{algpseudocode}
\usepackage{pgfplots}
\usepackage{multirow}
\usepackage{booktabs} 
\usepackage{subfig}
\pgfplotsset{compat=1.5}

\usepackage{amsmath}
\usepackage{amssymb}
\usepackage{dsfont}
\usepackage{bm}
\usepackage{xcolor,soul,colortbl}
\usepackage{array}
\usepackage{multirow}
\usepackage{caption}
\usepackage{dashrule}
\usepackage{graphicx}
\usepackage{breqn}
\usepackage{xspace}

\newcommand{\bh}{\mm z} % latent code
\newcommand{\gX}{\mathcal{X}^{gen}} % Dataset of generated images
\newcommand{\rX}{\mathcal{X}^{real}} % Dataset of real images
\newcommand{\hsm}{HSM} % hard sample mining with latent code optimization
\newcommand{\ds}{DS} % dataset smoothing
\newcommand{\bna}{BNA} %Batch norm adaptation

\newcommand{\clf}{C_{\theta}}
\newcommand{\clfbn}{C_{\theta,\omega}}
\newcommand{\ci}{C^{(i)}_{\theta}}
\newcommand{\BN}{BN}
\newcommand{\B}[0]{\mathcal B}

\newcommand{\R}[0]{\mathbb R}
\newcommand{\mm}[1]{\ensuremath{\bm{#1}}} %Matrix (upper case)
\newcolumntype{C}[1]{>{\centering\arraybackslash}p{#1}}	
\newcolumntype{L}[1]{>{\raggedright\arraybackslash}p{#1}}
\newcolumntype{R}[1]{>{\raggedleft\arraybackslash}p{#1}}

\newcommand{\bx}{\mm x}

\DeclareMathOperator*{\argmax}{arg\,max}

\makeatletter
\DeclareRobustCommand\onedot{\futurelet\@let@token\@onedot}
\def\@onedot{\ifx\@let@token.\else.\null\fi\xspace}

\def\eg{\emph{e.g}\onedot} 
\def\ie{\emph{i.e}\onedot}

\def\etal{\emph{et al}\onedot}
\makeatother
\title{This dataset does not exist: training models from generated images}

\name{Victor Besnier\textsuperscript{* 1,2}\thanks{*Victor Besnier worked on this project as an intern at Valeo.ai, Paris.}, Himalaya Jain\textsuperscript{1}, Andrei Bursuc\textsuperscript{1}, Matthieu Cord\textsuperscript{1,2}, and, Patrick P\'erez\textsuperscript{1}}
\address{
\textsuperscript{1}Valeo.ai, Paris, France\\
\textsuperscript{2}Sorbonne University, Paris, France\\
}

\begin{document}

\maketitle
\begin{abstract}
Current generative networks are increasingly proficient in generating high-resolution realistic images. 
These generative networks, especially the conditional ones, 
can potentially become a great tool for providing new image datasets. 
This naturally brings the question: \textit{Can we train a classifier only on the generated data?}
This potential availability of nearly unlimited amounts of training data challenges standard practices for training machine learning models, which have been crafted across the years for limited and fixed size datasets.
In this work we investigate this question and its related challenges. 
We identify ways to improve significantly the performance over naive training on randomly generated images with regular heuristics. 
We propose three standalone techniques that can be applied at different stages of the pipeline, \ie, data generation, training on generated data, and deploying on real data. We evaluate our proposed approaches on a subset of the ImageNet dataset and show encouraging results compared to classifiers trained on real images.
\end{abstract}

\section{Introduction}
% \vspace{-5pt}
In recent years, generative networks have made great progress. The state-of-the-art generative adversarial networks (GANs) can generate diverse high-resolution images almost indistinguishable from real images. Among others one of the main objectives of learning generative models is to get an unbounded supply of data for training machine learning models, in particular deep neural networks. 
This could significantly reduce expenses and work required for data acquisition or could be helpful when the original data cannot be shared, \eg~due to privacy. Moreover, storing a generator network is often lighter than storing the entire dataset, for example BigGAN~\cite{Brock2018} takes 225MB while ImageNet~\cite{russakovsky2015imagenet} takes $\sim$150GB.

Using GANs to generate abundant and diverse training data 
has been a longstanding motivation for research on GANs. However until recently, the obvious low quality of the generated images (\eg~low resolution, many artefacts) discouraged any attempts of training classifiers with such images only. Researchers focused mostly on improving the photorealism and diversity of the images thanks to the newly introduced metrics to measure these attributes, namely Inception Score (IS)~\cite{salimans2016improved} and Frechet Inception Distance (FID)~\cite{heusel2017gans}. Some tried using synthetic images for augmenting the existing set of real images to boost up performance \cite{tran2017bayesian, lim2018doping, mariani2018bagan}.

The current state-of-the-art for conditional image generator is BigGAN~\cite{Brock2018}. BigGAN has received lot of attention as it can generate high resolution, photorealistic, diverse images for all 1,000 semantic classes of ImageNet. Motivated by the outstanding IS and FID scores of BigGAN, Ravuri and Vinyals~\cite{Ravuri2019a} train a classifier over images generated from BigGAN and test it on the validation set of ImageNet. They find that
there is a significant drop in accuracy, $27.9\%$, compared to the classifier trained on the real images from training set of ImageNet\footnote{We note that the training set of ImageNet is very large, $\sim$1,300 images per class,  making it very challenging for a classifier trained on the same amount of generated data to outperform one trained on ImageNet.}, concluding that IS and FID are not suitable metrics for this task.

\begin{figure}[t]
    \centering
    \includegraphics{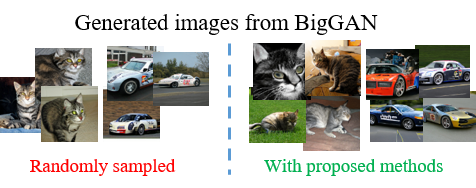}
    \caption{BigGAN can potentially generate diverse and photorealistic images. However, with random sampling, the diversity is much lower than a real image dataset of the same size. The proposed methods increase diversity and effective number of samples, leading to better performance for a classifier trained on generated images.
    }
    \label{fig:teaser}
\end{figure}

In this work, we address this challenging question of drop in performance when deep classifiers are trained on synthetic generated data and tested on real images. 
We estimate that the drop is probably due to the fact that the generated images, when sampled randomly, are less diverse than  those in the real dataset. However, since we have the generator we should be no longer limited in the amount of training samples we can have nor confined to randomly sampling from the generator (Fig.~\ref{fig:teaser}).
To overcome this drop in performance, we propose in this paper three learning strategies:
(1) a method to generate more informative data from the generator via latent code optimization; (2) a pragmatic strategy to use continuous sampling from a generator while training and hence, to increase diversity of the data; (3) a test time adaptation method to improve accuracy. Our three contributions may be combined and we will evaluate all the gains with BigGAN trained on ImagNet.

\section{Related Work}
\textbf{Improvement of GANs.} 
GANs are notoriously unstable and difficult to train, requiring several specific architectural choices and hyperparameter tuning. Several works have improved stability of GAN training via optimization~\cite{arjovsky2017wasserstein, Gulrajani2017} or architectural means~\cite{miyato2018spectral}.
BigGAN~\cite{Brock2018} is a pinnacle of both scientific and engineering expertise
accumulated across years of dealing with GANs. BigGAN is a class-conditional GAN whose training benefits from large batch sizes, high resolution images, 
and skip-connections for the latent code in the generator.
The quality of the generated images is improved through sampling from Gaussian noise with truncation.
Given the quality of BigGAN samples, we consider it to study training over synthetic images.\footnote{Given the cost of training BigGAN (24-48 hours on a Google TPU v3 Pod with 128 core for the smallest model with $128 \times 128$ images), we do not train it ourselves and use the pretrained model provided by the authors.}

\textbf{Latent code optimization.}
Exploring a GAN's latent space through iterative optimization has been proposed to generate 
more memorable images~\cite{goetschalckx2019ganalyze}, different geometric and photometric transformations of a given image~\cite{jahanian2019steerability} or faces with specific attributes~\cite{shen2019interpreting}.
The focus of these works is to generate realistic images. However this does not ensure that a classifier trained on such images would do well. Our latent code optimization aims to improve the performance of the image classification network and we use it for mining helpful codes for learning.

\textbf{Training over generated data.}
In the quest for more relevant metrics for evaluating GANs, recent studies started analyzing the performance of classifiers trained on synthetic images~\cite{Shmelkov2018, Ravuri2019, Ravuri2019a}. Shmelkov \etal~\cite{Shmelkov2018} show that all popular GANs consistently under-perform by a significant margin against classifiers trained on real images. Ravuri and Vinyals~\cite{Ravuri2019a, Ravuri2019} show that even for highly photorealistic images, \eg~from BigGAN~\cite{Brock2018}, synthetic classifiers lag behind real ones. Minor gains ($ + 0.2\%$) can be obtained by mixing real and synthetic images in specific settings.
In this work, we show that nontrivial distribution information learned by recent GAN architectures can be mined and leveraged towards competitive classifiers trained with supervision from synthetic data only.

\section{Approach}\label{sec:approach}

\begin{figure*}[t]
    \centering
    \includegraphics[width=0.80\textwidth]{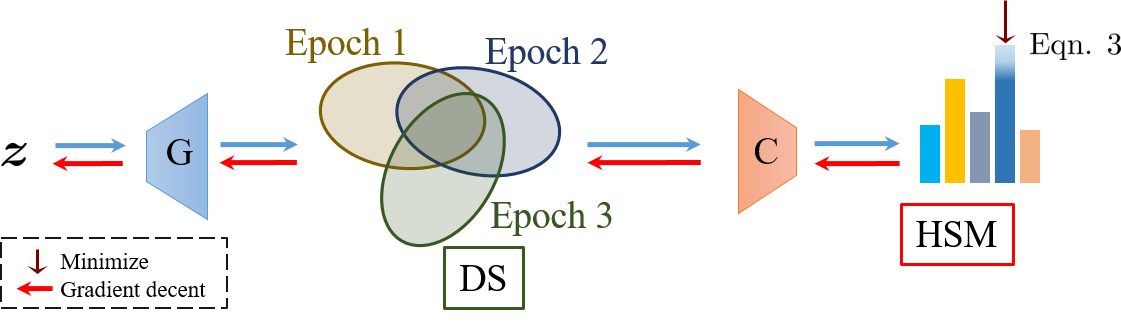}
    \caption{\textbf{Overview of the proposed \hsm~and \ds~methods for training.}
    In HSM, we minimize the maximum of logit over $\bh$. In \ds, at every epoch we replace some part of the data with new samples. Note that the methods work at different stages of the full pipeline and can be applied independently.}
    \label{fig:approach}
\end{figure*}

The aim of this work is to propose methods to use off-the-shelf pre-trained \textit{latent variable generative models} for training deep networks. Specifically we consider the set-up where, given only a pre-trained class-conditional GAN for image generation, one wants to train an image classifier with only images generated by this GAN. 

Now to describe our approaches, we first introduce the notation. Let $G$ be a pre-trained class-conditional generative network, which 
maps a latent code $\bh$ given a class label $y$ to an image $\bx$ \ie, $\bx = G(\bh, y)$. The generator is assumed of sufficient quality for the image $\bx$ to be semantically classified as $y$ by a human. 
We are interested in generating a dataset $\gX$ of images with labels which can be then used to train a classifier $\clf$ with parameters $\theta$. A classifier takes an image as input and gives scores for each class,
\begin{equation}
    \mm s_{\bx} = \clf(\bx) \in \R^K,
\end{equation}
where $K$ is number of classes. 
We train the classifier by iteratively minimizing the cross-entropy loss over the training set. The loss on one image is defined as
$- \log (\mm s_{\bx}[k]),$
where $k$ is the true class of $\bx$.

Note that when training the classifier on generated images, we do not update the generator $G$. As noticed in~\cite{Ravuri2019}, we also observe that if we simply sample the dataset $\gX$ from $G$ to train $\clf$, there is a significant drop in accuracy compared to 
% if $\clf$ was trained on $\rX$
training $\clf$ on a real dataset $\rX$. 
Reducing this performance gap is our aim. To this end,  
we propose three distinct and independent methods, as described below.

\subsection{Hard Sample Mining (\hsm) with latent code optimization}\label{sec:hsm}
Hard mining is a popular method in metric learning. The key intuition is to use harder or more informative samples from the training data to train the model. Such hard samples are interesting because they lead to increased robustness and better accuracy.

Now when training from generated images, we can use the generator to produce more informative images rather than drawing mere samples.
%ing it randomly. 
The more informative or harder images 
% depend on, and 
are specific to the current state of the classifier. 
Let $\ci$ be the classifier at an iteration $i$ during the training. The hard images for $\ci$ would be the images which are difficult to classify correctly by $\ci$,
\ie, with low classification score for the true class. 
For generating such an image for a given class $y$, we propose to optimize iteratively an original, random latent code $\bh$ to minimize the score of the class predicted by $\ci$ (not of the true class, as will be discussed later) for new image $G(\bh_{\mathrm{new}}, y)$ \ie,
\begin{equation}
    k^* = \argmax \ci\big(G(\bh, y)\big),
\label{eq_s}
\end{equation}
\begin{equation}
    \bh_{j+1} = \bh_j - \eta \nabla_{\bh_j} \mm s_{\bh_j, y}[k^*],
\label{eq_h}
\end{equation}
where $\bh_0=\bh$ and, $\mm s_{\bh_j, y}$ stands for $C_{\theta}^{(i)}\big( G(\bh_j,y)\big)$, hence $\mm s_{\bh_j, y}[k^*]$ is the score of class $k^*$ with current classifier applied to image $G(\bh_j, y)$ (Fig.~\ref{fig:approach}).
Here all the parameters of $G$ and $\ci$ are fixed and only the input of $G$, \ie,
the latent code, is optimized by back-propagating through $G$ and $\ci$. The optimized latent code will lead to a difficult image for $\ci$. 
However this image is not necessarily a good or valid sample. Since we optimize only for sample difficulty, the latent code can change too strongly across iterative updates and it may no longer follow the Gaussian prior expected by the generator. 
We temper this drift and preserve the prior across updates by scaling $\bh_j$ to have the same $\ell_2$-norm as $\bh_0$: $\bh_{j} \leftarrow\bh_{j}\frac{\|\bh_0\|}{\|\bh_{j}\|}$.

Note that in the objective above we do not minimize the score of the correct class $y$, and
we minimize instead the score of 
the highest confidence class (regardless of the true label). If the prediction is different from the true class, the image is already difficult and thus our objective will lead to an easier sample. 
This avoids generating overly hard samples, unlikely to be learned by the classifier.

\begin{figure}[t]
    \centering
    \subfloat{\scalebox{0.3}{\includegraphics{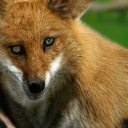}}} \hspace{.1pt}
    \subfloat{\scalebox{0.3}{\includegraphics{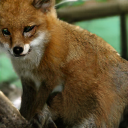}}} \hspace{.1pt}
    \subfloat{\scalebox{0.3}{\includegraphics{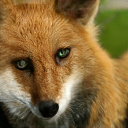}}} \hspace{.1pt}
    \subfloat{\scalebox{0.3}{\includegraphics{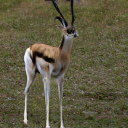}}} \hspace{.1pt}
    \subfloat{\scalebox{0.3}{\includegraphics{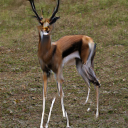}}} \hspace{.1pt}
    \subfloat{\scalebox{0.3}{\includegraphics{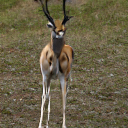}}} \\ \vspace{-7pt}
    \subfloat{\scalebox{0.3}{\includegraphics{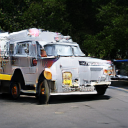}}} \hspace{.1pt}
    \subfloat{\scalebox{0.3}{\includegraphics{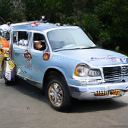}}} \hspace{.1pt}
    \subfloat{\scalebox{0.3}{\includegraphics{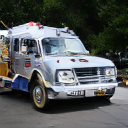}}} \hspace{.1pt}
    \subfloat{\scalebox{0.3}{\includegraphics{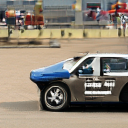}}} \hspace{.1pt}
    \subfloat{\scalebox{0.3}{\includegraphics{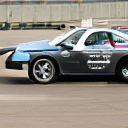}}} \hspace{.1pt}
    \subfloat{\scalebox{0.3}{\includegraphics{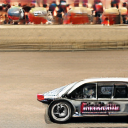}}} \hspace{.1pt}
    \caption{\textbf{Effect of applying HSM at different iterations during classifier training.} The first image of each category is sampled randomly. The other two images are generated from HSM-computed codes at two different steps during training. The difference between the images shows that the effect of HSM is specific to the classifier. 
    }
    \label{fig:hsm_exp}
\end{figure}

\textbf{Collecting a dataset with \hsm}. 
Since the difficulty of a sample is specific the performance of the model at a given iteration, in order to build a relevant and diverse dataset we should aggregate such samples from 
various intermediate stages of training.
To this end, we propose a simple algorithm to use \hsm~for collecting a fixed size dataset $\gX$ of $N$ samples within $K$ semantic classes as given in Alg. \ref{alg:fs}.  The underlying idea is to initialize the dataset with $M<N$ samples,
train a classifier on this data, then add to the dataset $M$ new samples collected with \hsm~for the classifier. We keep alternating between training the classifier and collecting $M$ new samples with \hsm~until the dataset $\gX$ reaches 
$N$ items. In Fig.~\ref{fig:hsm_exp}, we provide a few qualitative examples to illustrate the impact of HSM over the generated images. Once the gathering of the entire set $\gX$ is complete, we train a new, final classifier on it.

\setlength{\textfloatsep}{5pt}

\setlength{\textfloatsep}{5pt}
\begin{algorithm}[tbp]
    \caption{Collecting fixed size dataset with \hsm}
    \label{alg:fs}
    \begin{algorithmic}[1]
    \small
        \State Input: $G$, $C$, $N$, $K$
        \Comment{$N$: dataset size, $K$: \#classes}
        \State Output: $\gX$
        %\State $\gX \leftarrow \{\}$
        \Function{FillDataset}{$M$, {\sc withHSM}, $C$}
        \State $\mathcal{X} \leftarrow \{\}$
        \For {$k=1\cdots K$}
            \For{$m=1\cdots M/K$}
                \State $\bh \leftarrow \mathcal{N}(0, 1)$
                \If{{\sc withHSM}}
                    \State $\bh \leftarrow$ HSM($\bh, k, G, C$)\;  \Comment\parbox[t]{.35\linewidth}{mine hard code for sample $\bh$ for class $k$
                    }
                    
                \EndIf
                \State $\bx \leftarrow G(\bh, k)$
                \State $\mathcal{X}.${\sc Append}($\bx$)
            \EndFor
        \EndFor
        \EndFunction
        \State $i \leftarrow 0$
        \State $C_i \leftarrow C$
        \State $\gX \leftarrow$ \Call{FillDataset}{$M$, {\sc False}, {\sc None}}
        \Comment{$M<N$}
        \For {$i=1\cdots N/M$}
            \State  $C_i \leftarrow$ {\sc TRAIN}($C_{i-1}$, $\gX$) \Comment\parbox[t]{.4\linewidth}{train $C_{i-1}$ for an epoch
            }
            \State $\mathcal{X}^{hsm} \leftarrow$ \Call{FillDataset}{$M$, {\sc True}, $C_i$}
            \State $\gX.${\sc Append}($\mathcal{X}^{hsm}$)
        \EndFor
        \State {\sc Return} $\gX$
    \end{algorithmic}
\end{algorithm}

\begin{algorithm}[tbp]
    \caption{Classifier training with \ds~and \hsm}
    \label{alg:ds}
    \begin{algorithmic}[1]
    \small
        \State Input: $G$, $C$, $K$, $r$, $N$ \Comment{$N$: \#samples per epoch 
        }
        \State Output: $C$
        \State $i \leftarrow 0$
        \State $C_i \leftarrow C$
        \State $\gX_i \leftarrow$ \Call{FillDataset}{$N$, {\sc False}, {\sc None}}
        \For{$i=1\cdots I$}
            \State $C_i \leftarrow$ {\sc TRAIN}($C_{i-1}$, $\gX_i$)
            \State $\mathcal{X}^{hsm} \leftarrow$ \Call{FillDataset}{$rN/2$, {\sc True}, $C_i$} 
            \State $\mathcal{X}^{new} \leftarrow$\Call{FillDataset}{$rN/2$, {\sc False}, {\sc None}}
            \State $\gX_i \leftarrow$ {\sc Subset}( $\gX_i$) of size $(1-r)N$ 
            \State $\gX_{i+1} \leftarrow \gX_i \cup \mathcal{X}^{hsm} \cup \mathcal{X}^{new}$ 
        \EndFor
        \State {\sc return} $C_I$
    \end{algorithmic}
\end{algorithm}

\subsection{Dataset Smoothing (\ds)}\label{sec:ds}
As previously discussed, the models trained on generated data $\gX$ have lower performance compared to $\rX$ when using sets of equal sizes, \ie, $\lvert\gX\rvert=\lvert\rX\rvert$. 
This is caused by the lower visual diversity of the generated data in comparison to real data. 
We can afford to forego the dataset size constraint and augment the diversity through additional examples, since we have a generator that can practically generate unlimited training data.
Yet storing such a large dataset is not practical either. So, it is more pragmatic to generate the images online while training. 
On the downside, generating data perpetually will increase the training time significantly due to the non-negligible cost of generating images.
Moreover, training on this continuously changing data causes training instability, particularly in the first epochs.
This is probably due to differences in statistics of the generated data across training batches.

To mitigate these drawbacks, we propose our approach of \textit{dataset smoothing}. The key idea is to progressively change the dataset by only partially replacing the generated training data with new samples every epoch, aiming for a diverse but gradually changing dataset (Fig.~\ref{fig:approach}). In detail, we start with a generated dataset $\gX$, train the classifier on it. Then, we replace a fraction $r$ of $\gX$ with new randomly sampled images. This can be combined with \hsm~by applying it on some of the new images. The approach with \hsm~is summarized in Alg. \ref{alg:ds}. 
With the smooth changing of the dataset, we effectively have the diversity as in continuous sampling, while preserving a more stable and much faster training.

\begin{table*}
 \centering
  \begin{tabular}{ccccccccccccc} \toprule
     & \multicolumn{4}{c}{Fixed dataset} && \multicolumn{4}{c}{Continuous sampling} & ~& Long training &real images \\ \cmidrule{2-5}\cmidrule{7-10}\cmidrule{12-12}
    HSM & - & \checkmark & - & \checkmark & ~& - & \checkmark & - &\checkmark & ~& - & \\ 
    DS  & - & - & - & - & ~ & \checkmark & \checkmark & \checkmark & \checkmark & ~& \checkmark & \\
    BNA & - & - & \checkmark & \checkmark & ~ & - & - & \checkmark & \checkmark & ~& \checkmark & \\ \midrule
    Top-1 Accuracy & 70.6 & 73.6 & 76.4 & 78.2 & ~& 78.6 & 81.8 & 84.2 & \textbf{85.6} & ~& \textbf{88.8} & 88.4 \\ \bottomrule
\end{tabular}
  \caption{\textbf{Results for ImageNet-10 real test images.} Performance of classifiers trained on generated images with all combinations of the proposed methods. Each classifier is trained for $150$ epochs (except \textit{Long training}, where we let \ds~run for 1500 epochs) over a set of $N=13\text{K}$ images; in case of \emph{continuous sampling} we replace $50\%$ (\ie, $6,500$) of the images every epoch, while \emph{fixed dataset} is the usual setup where no images are replaced during training. In all setups we use $N$ images per epoch. First column, without applying any of the proposed methods, is the baseline. Each of the proposed methods individually shows improvement over the baseline. The combination of the methods further improves the results.
}
  \label{tab:10classes}
\end{table*}

\subsection{Batch Norm statistics Adaptation (\bna)}\label{sec:bna}
Batch Normalization (\BN)~\cite{ioffe2015batch} is a prominent architectural innovation in deep neural networks 
widely used for robustifying and accelerating training. 
\BN~stabilizes the distribution of hidden activations over a mini-batch during training by transforming the output of a layer 
to have zero mean and unit variance.
BN layers first set the mean and variances of activations,
and subsequently scale and shift them via a set of trainable parameters 
to maintain expressivity.

Formally, consider a mini-batch $\B$ of size $m$ and the 
respective activations $h_i^{(k)} \in \R, i=\{1 \dots m\}$ of a neuron $k$ for each sample in this mini-batch.
We omit $k$ for clarity.
\BN~transforms $h_i$ as follows:
\begin{equation}
\label{eq:bn_def}
\hat{h}_i=\gamma\frac{h_i-\mu_{\mathcal{B}}}{\sqrt{\sigma_{\mathcal{B}}^2+ \varepsilon}} + \beta,
\end{equation}
where $\mu_{\mathcal{B}}=\frac{1}{m}\sum_{i=1}^m h_i$ and $\sigma_{\mathcal{B}}^2=\frac{1}{m}\sum_{i=1}^m (h_i - \mu_{\mathcal{B}})^2$ are the mean and variance of mini-batch $\mathcal{B}$ for the considered neuron, $\varepsilon > 0$ is a small constant
, and $(\gamma, \beta) \subset \theta$ are learnable parameters.

During training, \BN~normalizes each mini-batch using its 
respective statistics $\mu_{\mathcal{B}}$ and $\sigma^2_{\mathcal{B}}$ computed on the fly. At the same time \BN~estimates the mean and variance of the activations from the whole training set, denoted by $\bar{\mu}$ and $\bar{\sigma}^2$, through exponential running averages with sub-unitary update factor $\alpha$. Formally, at training iteration $t$, the running mean and variances are given by:
\begin{align}
\label{eq:bn_running_stats}
\bar{\mu}^t &= \alpha \mu^{t-1}_{\B}  + (1-\alpha)\bar{\mu}^{t-1} \\ \nonumber
(\bar{\sigma}^2)^t &= \alpha (\sigma^2_{\B})^{t-1}  + (1-\alpha)(\bar{\sigma}^2)^{t-1}.
\end{align}
We denote the set of estimated statistics for all \BN~layers in our classifier network as $\omega=\{ \bar{\mu}, \bar{\sigma}^2 \}$. A neural network with \BN~layers is parameterized by two types of parameters: $\theta$, learned by backpropagation and gradient descent, and $\omega$, computed from feature activation statistics. We denote our classifier using \BN~as $C_{\theta, \omega}$.

During testing, \BN~uses the estimated mean and variances for normalizing input activations from test samples,
enforcing distributions of activations similar with training.
\BN~assumes that both train and test images are sampled from similar 
distributions and the expectations of the 
activation values
from the two sets are close to each other. 
Thus \BN~can be sensitive to differences between train and test data, which occur for instance in multi-domain training~\cite{rebuffi2017learning}, domain adaptation~\cite{chang2019domain}.
Albeit BigGAN generates photorealistic samples, as we train exclusively on synthetic images and test on real images, the classifier $\clfbn$ might be sensitive to this small domain gap.

In order to mitigate this, we leverage \BN~layers for a simple unsupervised domain adaptation of $\clfbn$ from synthetic to real images. 
We adapt the estimated \BN~statistics for real images through 
a forward pass on unlabeled images from the training set, without backpropagation, 
as the running averages for $\bar{\mu}$ and $\bar{\sigma}^2$ are updated on the fly. 
In detail we compute $\omega^{real}$ from real images with Eq.~\ref{eq:bn_running_stats}, while keeping the parameters $\theta$ intact.
We refer to this method as BatchNorm statistics Adaptation (BNA).

\section{Experiments}

In this section we evaluate the three proposed approaches to learn with generated data. We first outline the datasets we use and describe the evaluation method. Then, we give the details of experimental setup followed by the experimental results and their analysis.

\textbf{Dataset.} As we train only on generated images, we use the real dataset just for testing and not for training. Since we use BigGAN as generator, which is trained to generate ImageNet-like images~\cite{russakovsky2015imagenet}, we use ImageNet for evaluation. ImageNet is a large dataset of 1.3M labeled images, with 1K semantic classes. As the dataset is very large, we select 10 classes to evaluate our approaches with reasonable computation time and resources. We refer this subset of 10 classes as  ImageNet-10. We select the classes of ImageNet-10 to mimick the ones from CIFAR-10 dataset~\cite{krizhevsky2009learning}\footnote{CIFAR-10 classes are broad categories, while ImageNet ones are fine-grain. Thus we select representative classes from ImageNet manually, \eg~\emph{labrador} for \emph{dog}. In place of CIFAR-10 classes with no equivalent (\emph{frog}, \emph{horse}) we use \emph{red fox} and \emph{balloon} reported as top and respectively worst performing classes in the experiments from~\cite{Ravuri2019a}.}.
The 10 classes are: \emph{aircraft carrier}, \emph{balloon}, \emph{fire truck}, \emph{gazelle}, \emph{goldfish}, \emph{labrador}, \emph{red fox}, \emph{redshank},  \emph{race car}, \emph{tabby cat}. 
We use as test set, images from the validation set of ImageNet for the selected 10 classes, with 50 images per class.
Finally, we use real images from the train set of ImageNet-10 to train a classifier on real images to compare against.

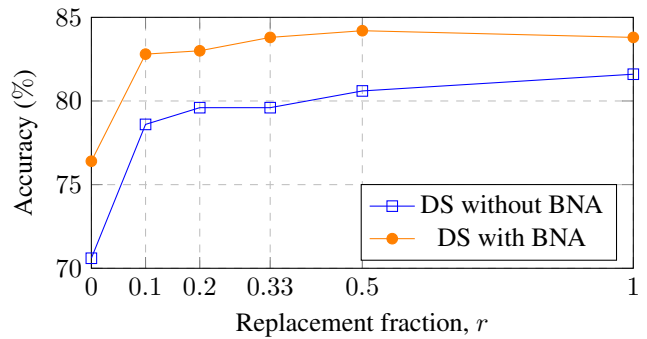
\begin{figure}[tbp]
    \centering
    \begin{tikzpicture}[scale=1]
    \begin{axis}[
        width=250pt,height=140pt,
        title={}, 
        xlabel={Replacement fraction, $r$}, ylabel={Accuracy ($\%$)},
        xmin=0, xmax=1, ymin=70, ymax=85,
        xtick={0,0.1,0.2,0.33,0.5,1},
        ytick={70,75,80,85},
        legend pos=south east,
        %ymajorgrids=true,
        grid=both, grid style=dashed,
        ]
        \addplot[color=blue, mark=square,]
        coordinates {(0.0,70.6)(0.1,78.6)(0.2,79.6)(0.33,79.6)(0.5,80.6)(1,81.6)};
        \addplot[color=orange, mark=*,]
        coordinates {(0.0,76.4)(0.1,82.8)(0.2,83.0)(0.33,83.8)(0.5,84.2)(1,83.8)};
        \legend{\ds~without \bna, \ds~with \bna}
    \end{axis}
    \end{tikzpicture}
    \caption{\textbf{Effect of replacement fraction $r$ in \ds.} Classification accuracy using \ds~on real images with varying $r$, \ie, fraction of the dataset being replaced with new images every epoch. The figure shows plots for \ds~with and without BNA.
    }
    \label{fig:recycle}
\end{figure}

\textbf{Experimental setup.}
We use as evaluation metric the Top-1 accuracy on real images, \ie, the percent of test images classified correctly.
For our experiments, we use as generator a pre-trained BigGAN~\cite{Brock2018} that produces $128\times128$ images.\footnote{https://github.com/huggingface/pytorch-pretrained-BigGAN} For all the experiments we use ResNet-18 as classifier. We use SGD optimizer with $lr=0.1$, momentum, learning rate decay and weight decay $1e-4$.

\textbf{Results.} We first assess the impact of our contributions individually and in combination (Table~\ref{tab:10classes}). 
We use as baseline a classifier trained on random $\gX$ without any of the proposed approaches. 
\hsm~brings gains from $1.4\%$ to $3.2\%$ whatever the combination. \ds~improves the accuracy by $7.4\%$ to $8.2\%$. With \bna, which is just a post-training adaptation on real data, scores get a boost of $3.8\%$ to $5.8\%$. Interestingly, all methods are complementary and any combination improves scores. Overall, by combining all three proposed methods, we outperform the baseline by $15\%$. By increasing the number of training iterations $10$ times, we reach $88.8\%$ accuracy.

\textbf{Ablation studies.}
We study the impact of DS with continuous sampling, \ie, dynamically generating new samples every epoch.
We use as baseline a classifier trained over a fixed size dataset $\gX$ of $N$ generated images, with $\lvert\gX\rvert=\lvert\rX\rvert$. For DS, we train another classifier for which at every epoch we sample $r\times N$ new images and replace a part of the generated dataset with them. At each epoch we have a constant number of images $N$.
% $\lvert\gX\rvert=\lvert\rX\rvert$. 
In Fig.~\ref{fig:recycle}, we show the impact of the replacement fraction on classification accuracy. The fraction $r$ is linearly related to the amount of samples generated and time it requires during training. Thus, it is compelling to have lower $r$. 
In general, increasing $r$ improves accuracy, especially between \emph{no replacement} and $r=0.1$. The improvement diminishes quickly and almost saturates after $r=0.5$ at the expense of longer training time.
In the same setup, we can see that BNA systematically improves accuracy with up to $+5.8\%$. Interestingly, using \bna~over DS with $r=0.5$ has slightly better accuracy than replacing all ($r=1$) which moreover takes longer to train.

\section{Conclusions}
In this work we address the task of training a classifier using solely generated data. 
We show that for practically unlimited amounts of data, several standard practices in training models could be revisited. We propose three standalone contributions to this effect and show their benefits both when used individually or combined. The limitations of this study are related to the reduced number of considered classes due to computation constraints and due to high variability in the quality of BigGAN images across classes. Given the fast pace of progress in GAN literature, steady improvements in the quality of the generated images are foreseeable. We expect that the same questions addressed here will still arise when training over generated data from future GANs and our proposed approaches are readily applicable in such context.

% \section{REFERENCES}
\label{sec:refs}
\bibliographystyle{IEEEbib}
\bibliography{citation}

\end{document}